\documentclass[11pt,a4paper]{article}
\usepackage[hyperref]{acl2019}
\usepackage{times}
\usepackage{latexsym}
\usepackage{amsmath}
\usepackage{amsfonts}
\usepackage{booktabs}
\usepackage{multirow}

\usepackage{url}

\aclfinalcopy

\title{A Good Sample is Hard to Find: Noise Injection Sampling and 
    Self-Training for Neural Language Generation Models}

\author{Chris Kedzie \\
  Columbia University \\
  Department of Computer Science \\
  \texttt{kedzie@cs.columbia.edu} \\\And
  Kathleen McKeown \\
  Columbia University \\
  Department of Computer Science \\
  \texttt{kathy@cs.columbia.edu} \\}

\date{}

\begin{document}
\maketitle
\begin{abstract}
    Deep neural networks (DNN) are quickly becoming the 
    de facto standard modeling method for many natural
    language generation (NLG) tasks. In order 
    for such models to truly be useful, they must be capable
    of correctly generating utterances for novel meaning
    representations (MRs) at test time. 
    In practice, even sophisticated DNNs with various forms of semantic
    control frequently fail to generate utterances faithful to the 
    input MR. 
                                         In this paper, we propose an
   architecture agnostic 
    self-training method to sample novel  MR/text utterance
    pairs to augment the original training data.
Remarkably, after training on the augmented data,
    even simple encoder-decoder models with greedy decoding are 
    capable of generating semantically correct utterances that are as
    good as state-of-the-art outputs in both automatic and human 
    evaluations of quality.

\end{abstract}

\newcommand{\gen}{\ensuremath{p}}
\newcommand{\basegen}{\ensuremath{p_0}}
\newcommand{\auggen}{\ensuremath{p_1}}
\newcommand{\ruleclf}{\ensuremath{q_{\tiny \mathfrak{R}}}}
\newcommand{\learnedclf}{\ensuremath{q_{\phi}}}
\newcommand{\clf}{\ensuremath{q}}
\newcommand{\mrx}{\ensuremath{x}}
\newcommand{\descy}{\ensuremath{y}}
\newcommand{\sampley}{\ensuremath{\tilde{\descy}}}
\newcommand{\samplex}{\ensuremath{\tilde{\mrx}}}
\newcommand{\modparams}{\ensuremath{\theta}}
\newcommand{\trdata}{\ensuremath{\mathcal{D}}}
\newcommand{\augdata}{\ensuremath{\mathcal{A}}}
\newcommand{\encstate}{\ensuremath{c}}
\newcommand{\decstate}{\ensuremath{h}}
\newcommand{\embsize}{\ensuremath{D}}
\newcommand{\npaddist}{\ensuremath{\basegen(\cdot|\mrx, \epsilon)}}

\section{Introduction}

Deep neural network (DNN) architectures have become the standard modeling 
method for a host
of language generation tasks. When data is plentiful, 
the sequence-to-sequence framework
proves to be incredibly adaptable to a variety of problem domains. Recent evaluations of end-to-end
trained DNNs for dialogue generation have shown that they 
are capable of learning very natural text realizations of formal
meaning representations (MRs),
i.e. dialogue acts (DAs) with slot-filler type attributes
(see \autoref{figure:introexample} for an
example).
In many cases,
they  beat rule and 
template based
systems on human and automatic measures of quality \cite{duvsek2019evaluating}.
  
\begin{figure}
\centering
\begin{tabular}{p{7cm}}
\toprule
$\begin{array}{ll}
    \textbf{Inform} &     \textrm{name[The Golden Curry]}\\
    &\textrm{near[The Six Bells]}\\
 &\textrm{\mbox{familyFriendly}[yes]}\end{array}$\\~\\

\textbf{Training Reference Utterance} \\
Near The Six Bells is a venue that is children friendly named The Golden Curry.\\
\bottomrule
\end{tabular}
\caption{Example MR for the Inform DA with example human reference utterance.}
\label{figure:introexample}
\end{figure}

However, this powerful generation capability comes with a cost; DNN language
models are notoriously difficult to control, often producing quite 
fluent but  semantically misleading outputs. 
In order 
    for such models to truly be useful, they must be capable
    of correctly generating utterances for novel MRs at test time. 
In practice,
   even with delexicalization \cite{duvsek2016sequence,juraskaslug2slug}, 
copy and coverage mechanisms \cite{elder2018e2e}, 
    and overgeneration plus reranking \cite{duvsek2016sequence,juraskaslug2slug}, DNN generators still produce 
errors \cite{duvsek2019evaluating}.

In this work, rather than develop more sophisticated DNN architectures
or ensembles,
we explore the use of  simpler DNNs  with self-training.
We train a bare-bones unidirectional neural encoder-decoder with attention
\cite{bahdanau2014neural}
 as our base model from which we sample novel utterances for MRs
not seen in the original training data. 
We obtain a diverse collection of samples using noise injection sampling 
\cite{cho2016noisy}.
Using an MR parser, we add novel utterances with valid MRs to the original
training data.
Retraining the model on the augmented data yields a language generator
that is more reliable than the sophisticated DNNs that have been recently
developed, in some cases reducing test set semantic errors to zero, without
sacrificing linguistic quality.

In this paper we make the following contributions. 1) We propose a general
method of data augmentation for natural language generation (NLG) problems 
using noise injection sampling and self-training. 2) We show a reduction
of attribute realization errors across several
dialog generation datasets, while achieving 
competitive automatic and human quality evaluation scores. 3) Finally, 
we show that 
these results hold even when the MR parser is noisy or we use fully 
lexicalized generation models.\footnote{Code and data for this paper can 
be found at \\
\url{https://github.com/kedz/noiseylg}}

\section{Datasets and Problem Defintion}
We ground our experiments in three recent dialogue generation datasets,
the E2E Challenge Dataset \cite{duvsek2019evaluating}, and the Laptops and 
TVs datasets \cite{wen2016multi}.
We only briefly review them here.
Each dataset consists of dialog act MRs paired with one or more 
reference utterances 
(see \autoref{figure:introexample} for an example from the E2E dataset).
The structure of each MR is relatively simple, consisting of the 
dialog act itself, (e.g. \textit{inform}, \textit{recommend}, 
\textit{compare}, etc.) and a variable number of attribute slots
which need to be realised in the utterance. All attribute values 
come from a closed vocabulary.
If an attribute is not present in the MR it should not be realized in the 
corresponding utterance. 

The three datasets also represent different training size conditions;
there are 42,061, 7,944, and  4,221 training examples in E2E, Laptops,
and TVs datasets respectively.

The NLG task for all three datasets 
is to produce an utterance for a given MR such that all attributes
in the MR are realized naturally and correctly.

\paragraph{E2E Model Input}
Following previous sequence-to-sequence approaches for the E2E dataset 
\cite{juraskaslug2slug},
we treat the MRs as a linear sequence of tokens
$\mrx = (\mrx_1, \ldots, \mrx_8)$ where each of the 8 positions represents
the value of a corresponding attribute.
If an attribute is not specified
in the MR we assign it an attribute specific \textit{n/a} token.
In the E2E dataset  there is only one dialog act type, \textit{Inform}, 
so we do not represent it in \mrx.

Prior work often delexicalizes the \textit{Name} and \textit{Near} 
attributes (i.e. replaces verbalizations of attribute values with a
placeholder token), which can later be replaced with the original
attribute value in a post-processing step. 
For example, the delexicalized version of the utterance in 
\autoref{figure:introexample}
would be ``Near NEAR is a venue that is children friendly named NAME.''
\textit{Name} and \textit{Near}  have a relatively large vocabulary of valid slot fillers, some of which are only seen  infrequently in the training data;
it can be difficult for fully lexicalized models to produce some of the 
rarer location names for these attributes. 

However, since delexicalization might be difficult or 
impossible in other domains, we implement both delexicalized and lexicalized
versions of the generation models on the E2E dataset to 
more fully evaluate the
self-training method.\footnote{
    Additional preprocessing details can be found in \autoref{app:preproc}.}

\paragraph{Laptops and TVs Model Inputs}
The Laptops and TVs datasets have a more diverse set of dialog acts
and can have repeated attributes (with different values)  in some cases, so we 
abandon our fixed length, fixed position encoding, and represent
each MR as a initial dialog act token and then a variable length sequence
of tokens for each of the specified attributes. 
The evaluation script for these datasets uses delexicalization
to evaluate attribute realization error, and so we use it here to be 
consistent with prior work, 
delexicalizing all possible attributes.
See \autoref{app:inputrep} for example input sequences for all datasets.

\section{Generation Model}
We treat the generation task as a sequence-to-sequence
 transduction problem, where we learn a probabilistic mapping $p$ from the 
linearized MR \mrx~
to a sequence of $N$ tokens $\descy = (\descy_1, \ldots, \descy_N)$
consituting the utterance. The model $p$ is implemented using
a two-layer unidirectional\footnote{In initial experiments, we found the unidirectional encoder to perform better than a bidirectional one.} encoder-decoder architecture with 
gated recurrent units \cite{cho2014learning} and
feed-forward
style attention \cite{bahdanau2014neural} as this is a 
canonical recurrent architecture for sequence-to-sequence modeling. 
We use 512 dimensions for all embeddings and GRU states.  

We fit the model parameters $\modparams$ by minimizing the negative log
likelihood of the training set $\trdata$, i.e. $\mathcal{L}(\modparams) = 
- \sum_{(\mrx, \descy) \in \trdata} \log p(\descy|\mrx; \modparams)$
using stochastic gradient descent. Going forward we omit $\modparams$ for
clarity.

\subsection{Generating from $p$}

\paragraph{Deterministic Decoding} 
Given an arbitrary MR \mrx, we can generate
an utterance \sampley~using greedy decoding, i.e. $\sampley_i =
\operatorname{arg\;max}_{\descy_i} p(\descy_i|\sampley_{<i}, \mrx)$.
To produce an ``$n$-best'' list of outputs, we can
also use beam decoding where $n$ candidate utterances are maintained
during each decoding step. Both greedy and beam decoding are known
to produce somewhat homogeneous outputs \cite{serban2016building}. 
Diversifying 
beam outputs often involves careful tuning of secondary search objectives
which trade off fluency \cite{li2015diversity}. 

Moreover, when training fully lexicalized models we found that we could 
often \emph{not} produce certain \textit{Name} and \textit{Near} attribute values. For example, we constructed a novel MR with \textit{near[Burger~King]}
and fed it into our base generator.
Even with an impractically large 
beam size of 512, we could not produce an utterance with ``Burger King'' in it.
This failure mode makes beam search a relatively unuseable method
for producing utterances for MRs under-represented in the training 
data.\footnote{The MR in this case had three attributes. \textit{near[Burger King]}
only occurs in size eight MRs in the training data.} 

To overcome this limitation we explored several sampling methods for generating these rarer utterances, namely ancestral and noise injection sampling.

\paragraph{Stochastic Decoding} 
Ancestral sampling (i.e. drawing a random token from
$p(Y_i|\descy_{<i}, \mrx)$ at each decoder step) is another option
for generating diverse outputs, but the outputs can be of lower fluency and 
coherence. Producing rare tokens often involves tuning a temperature 
parameter to flatten the output distribution, but this again can hurt fluency.

As an alternative, we can obtain diverse samples from greedy decoding 
by injecting Gaussian noise into the decoder hidden states following
\citet{cho2016noisy}.
Under our model, the probability $p(\descy_i|\descy_{<i}, \mrx) = f(\mrx,  h_{i})$ of the $i$-th token is a function of the encoder 
inputs \mrx~and the decoder hidden state $h_i \in \mathbb{R}^D$. 
When peforming noise injection sampling, 
we replace $h_i$ with a noisy state $\tilde{h}_i = h_i + \epsilon_i$
where $\epsilon_i$ is drawn from a $D$-dimensional Gaussian distribution
 $\mathcal{N}(0, \sigma^2_i I)$ with variance $\sigma^2_i = \frac{\sigma_0^2}{i}$. The base variance $\sigma_0^2$ is a a hyperparameter. 
 Effectively, the first few steps allow the decoder to 
 reach a novel hidden state space, while the gradually diminishing 
 noise allows the decoder to produce fluent outputs.

 \begin{table*}
    
    \begin{tabular}{p{\dimexpr \linewidth-2\tabcolsep}}
\toprule
\textbf{Input MR}  \\
\mbox{Inform(name[The~Cambridge~Blue], eatType[Restaurant],
customerRating[high], food[Italian])} \\
\textbf{Ancestral Sampling} \\
The Cambridge Blue is an Italian restaurant with a high customer rating. \\
The Cambridge Blue is an Italian restaurant with high ratings. \\
\textbf{*}Italian restaurant, the Cambridge Blue, has a high customer rating. ~~(Phrase fragments, not fluent.)\\
\textbf{Noise Injection Sampling} \\
The Cambridge Blue is a restaurant that serves Italian food. it has a high customer rating. \\
\textbf{*}The Cambridge Blue is a highly rated restaurant. ~~ (Drops \textit{food[Italian]}.)\\
\textbf{*}The Cambridge Blue is a restaurant located near the Bakers.~~ (Hallucinates \textit{near[The Bakers]}.)  \\
\bottomrule
\end{tabular}
\caption{Examples of ancestral sampling and noise injection sampling 
$(\sigma_0=1.0)$. \textbf{*} indicates output that is either not grammatical or is not correct with respect to the input MR. Text in parentheses explains the details of the error in either case.}
    \label{table:generationmethods}
 \end{table*}

 \paragraph{Generating Rare Values}
 Remarkably, the samples obtained by noise injection
 maintain fluency
 and valid syntactic structure. At the same time they often
 hallucinate or drop attributes. For example, see the last noise injection
 sample in \autoref{table:generationmethods} where the 
 attribute \textit{near[The~Bakers]} is hallucinated.
    This kind of error is actually useful because, as long as we have 
    a reliable MR parser, we can recover the MR of the sample
    and we now have a totally valid extra datapoint that we could use 
    for training.
 Syntactic errors of the kind produced by ancestral sampling
 (see the last ancetral sampling example in 
 \autoref{table:generationmethods}), on the 
 other hand, are not good to train on because they damage the 
 fluency of the decoder.

 Returning to the ``Burger King'' example, with noise injection
sampling we 
were able to produce over 10,000 novel instances of it. See \autoref{app:samples} 
for more samples.

\section{MR Parsing Model}

Given a novel utterance \sampley~sampled from $p$, we need to 
reliably parse the implied MR, i.e. $\samplex = \clf(\sampley)$, 
where \clf~is our parsing model. We have two things going for us in 
our experimental setting. First, even with noise injection sampling,
model outputs are fairly patterned, reducing the variability of the utterances
we need to parse in practice. 

Second, the MRs in this study are
flat lists of attributes that are somewhat independent of each other.
We only need to detect the presence of each attribute and its value.
For the Laptops and TVs datasets we also need to recover the dialog
act but these also are signaled by a fairly limited repertoire 
of cues, e.g. ``we recommend.'' Given this, we experiment with both hand crafted regular expression 
rules and learned classifiers to predict the value of
an attribute if present or that it is missing.

\paragraph{Rule-based parser \ruleclf} We design hand-crafted 
regular expression based rules to match for the presence of key phrases 
for each of the attributes and DAs in the datasets while also checking to
make sure that there is only one match per attribute.

To construct the rules, we look through both the training data references as 
well as the generation model outputs as this is what the rules will
be operating on in practice. For each lexicalized attribute (and DA) we 
develop a list of regular expressions such as,
\texttt{/is (family|kid|child) friendly/} $\Rightarrow$ \textit{familyFriendly[yes]}.
For the delexicalized attributes, we simply check for the presence 
of the placeholder token.

We design these rules to be high precision, as it is safer to miss out on 
more obscure varieties of utterance to avoid adding incorrectly parsed data 
points.
However,  in many cases the rules are also high recall as well. 
The average F-score on the E2E validation set is 0.93.

\paragraph{Classifier-based parser \learnedclf} 
It is perhaps too optimistic to believe we can construct reasonable rules
in all cases. Rule creation quickly becomes tedious and for more complex
MRs this would become a bottleneck. To address these concerns, we also 
study the feasibility of using learned classifiers to predict the presence
and value of the attributes. For each attribute in the E2E dataset,
we trained a separate convolutional neural network (CNN) classifier 
to predict the correct attribute value (or \textit{n/a} if the attribute is 
not present).
The CNN architecture follows that of \citet{kim2014convolutional} and is 
trained with 
gradient descent on the original training data. See \autoref{app:cnnclf} for
full architecture and training details.
The average E2E validation F-score is 0.94.

\section{Self-Training Methodology}
\label{section:selftrain}

Our approach to self-training is relatively straightforward and 
invariant
to the choices of whether or not to use delexicalization, and rule vs. 
classifier based
parser. There are minor differences depending on the dataset and we elaborate 
on
those below. There are three main steps to our self-training approach.
Starting with an initially empty augmented dataset \augdata, we 

\begin{enumerate}

    \item Train a base generator model \basegen~on the original training
        data \trdata.
    \item Repeat many times:
        \begin{enumerate}
            \item Sample a random MR $\mrx \sim \mathcal{X}$. 
            \item Sample $K$ utterances $\sampley^{(i)} \sim \basegen(\cdot|\mrx, \epsilon)$ 
            \item Parse MR, $\samplex^{(i)} = \clf(\sampley^{(i)})$,
            discarding any samples with invalid parses, and adding the 
            survivors to \augdata.
        \end{enumerate}
    \item Train a new generator \auggen~ on the combined dataset $\trdata \cup
       \augdata$.
\end{enumerate}
Steps 1 and 3 are identical, the generators \basegen~and \auggen~have the same 
architecture and training setup, ony the dataset, \trdata~vs. $\trdata \cup
\augdata$, is different. We now discuss step 2 in detail.

\paragraph{Step 2: E2E Dataset} 
To sample a novel MR with  $S$ attributes, we sample a combination of $S-1$ attributes
uniformly at random (always appending the \textit{name} attribute since every MR contains it).
 We 
then sample attribute values for each slot inversely proportional to their empirical frequency
in the training set so as to increase the likelihood of creating a novel
or under-represented MR.

After obtaining such a sample \mrx~we then perform noise injection sampling,
generating 200 samples $\sampley^{(i)} \sim 
\basegen(\cdot|\mrx, \epsilon^{(i)})$ 
in parallel and discarding all but the top 20
samples by average log likelihood according to \basegen.
We also discard any utterances that have previously been generated.

We then apply the parser to the sampled utterances, to obtain its
predicted MR, $\samplex^{(i)} = \clf(\sampley^{(i)})$. If using 
the rule based parser \ruleclf~and $\samplex = \emptyset$, i.e. the utterance
does not have a valid parse, we discard it.
Similarly, when using the classifier based parser, \learnedclf, if 
any attribute value is predicted with less than 50\% probability we discard 
it. All surviving $(\samplex^{(i)}, \sampley^{(i)})$ pairs are added to \augdata.
We repeat this process 25,000 times for each valid MR size $S$.
See \autoref{table:samplequal} for statistics on the total sample sizes 
after filtering.

\paragraph{Step 2: Laptops and TVs} On the Laptops and TVs dataset,
for each DA and legal number of attributes $S$ we draw $S$ random attributes
(modulo any required attributes like \textit{Name}; not all DAs require it).\footnote{A number of attributes $S$ is ``legal'' if we observe a DA instance with that 
many attributes in the original training data.}

We then perform noise injection sampling,
generating 200 samples $\sampley^{(i)} \sim 
\basegen(\cdot|\mrx, \epsilon^{(i)})$ under the same settings as the E2E
dataset. We repeat this process 25,000 times for each DA and DA size.
We obtain 373,468 and 33,478 additional samples for the Laptops 
and TVs datasets respectively.

\section{Experiments}

\subsection{E2E Self-Training}
 We train base generators \basegen~on the original training data \trdata, 
 with and without
delexicalizing the \textit{Name} and \textit{Near} attributes. 
We train for 500 epochs with gradient descent. We use a batch size of 128,
with a learning rate of 0.25, weight decay penalty of 0.0001, and a dropout 
probability of 0.25.
We select the best model iteration using validation
set BLEU score\footnote{We use the official shared task script to
compute automatic quality metrics on the E2E dataset.}.

Using the self-training method outlined in \autoref{section:selftrain},
we create augmented datasets using either \ruleclf~or
\learnedclf, which we refer to as 
$\augdata_{\ruleclf}$ and $\augdata_{\learnedclf}$ respectively 
(\learnedclf~is only in the delexicalized setting).

For both $\trdata \cup \augdata_{\ruleclf}$ and 
$\trdata \cup \augdata_{\learnedclf}$ we train new generators \auggen~using 
the same training setting as above (although we terminate training after 50 
epochs
because the models converge much faster with the additional data).

\begin{table}
\centering
\begin{tabular}{rcccc}
\toprule
Model  & BLEU & R.-L & MET. \\ \midrule
Slug   & 66.19 & 67.72 & 44.54  \\  
DANGNT & 59.90 & 66.34  & 43.46  \\
TUDA   & 56.57 & 66.14  & 45.29  \\
        \midrule
delex. \basegen~~~~~~greedy & 66.91 & 68.27 & 44.95 \\
                      beam & \textbf{67.13} &  \textbf{68.91} &  45.15  \\
        \auggen~\ruleclf~greedy   &  65.57 & 67.71 & 45.56  \\
                             beam & 66.28 &  68.08 &  \textbf{45.78}  \\
     \learnedclf~greedy & 63.76 &  67.31 &  44.94  \\
   beam   & 
 64.23 &  67.54 &  45.17  \\
\midrule
   lex. \basegen~~~~~~greedy & 60.35 &  64.51 & 41.82  \\
                      beam   & 61.81 &  65.83 & 42.69  \\
    \auggen~\ruleclf~greedy & 64.74 &  68.21 & 44.46  \\
       beam & 64.81 &  67.83 &    44.39  \\
\bottomrule
\end{tabular}
\caption{BLEU, ROUGE-L, and METEOR metrics on the E2E test set. Baseline methods all rely
on at least partial delexicalization, puting our lexicalized models at a relative disadvantage.}

\label{table:autoqual}
\end{table}

\paragraph{Results} \autoref{table:autoqual} shows the automatic quality measurements on
the E2E test set using BLEU, ROUGE-L, and METEOR.
We show results for both greedy and beam decoding with beam size 8
under \basegen~and
\auggen~models. We compare our models to the best sequence-to-sequence DNN
model, Slug \cite{juraskaslug2slug}, the best grammar rule based model, 
DANGNT \cite{nguyen2018structurebased},
and the best template based model, TUDA \cite{puzikov2018e2e}, as determined during 
the shared task evaluation \cite{duvsek2019evaluating}.

\begin{table*}
\setlength{\tabcolsep}{5pt}
\center
  \begin{tabular}{rrrr ccccc ccc ccc}
    \toprule
 \multicolumn{4}{c}{
\multirow{2}{*}{
Model} }
       & \multirow{2}{*}{Name} & \multirow{2}{*}{Near}    
            &  Family  & 
           \multirow{2}{*}{Area}    & Customer & \multirow{2}{*}{Food} 
      & Price & Eat & \multirow{2}{*}{All} \\
  & & & &  &  & Friendly & 
        & Rating &  & Range & Type &  \\
\midrule
\multicolumn{4}{r}{Slug}  
                    & 0 & 0 & 6  & 1  & 6  &  10  & 35  & 9  & 67 \\ 
\multicolumn{4}{r}{DANGNT}
                    & 0 & 0 & 18  & 0  & 0  & 0  & 0  & 58  & 76 \\ 
\multicolumn{4}{r}{TUDA}  
                    & 0 & 0 & 0  & 0  & 0  & 0  & 0  & 0    & \textbf{0} \\
\midrule
delex. & \basegen & & greedy 
                    & 0   & 0   & 23 & 23 & 16 & 26 & 27 & 0 & 115 \\ 
& & & beam          & 0   & 0   & 60 & 3  & 9  & 3  & 8  & 0 & 83  \\ 
& \auggen & \ruleclf & greedy 
                    & 0   & 0   & 0  & 0  & 0  & 0  & 0  & 0 & \textbf{0} \\ 
 & & & beam         & 0   & 0   & 0  & 0  & 0  & 0  & 0  & 0 & \textbf{0} \\
 & & \learnedclf & greedy 
                    & 0   & 0   & 1  & 0  & 8  & 1  & 9  & 0 & 19 \\
 & &  & beam        & 0   & 0   & 0  & 0  & 3  & 0  & 0  & 0 & 3 \\
\midrule
lex. & \basegen & & greedy 
                    & 145 & 141 & 14 & 15 & 2   & 14 & 2  & 0 & 333 \\
 & & & beam         & 155 & 124 & 62 & 0  & 0   & 0  & 0  & 0 & 341 \\ 
 & \auggen & \ruleclf  & greedy 
                    & 0   & 0   & 2  & 0  & 0  & 125 & 0  & 0 & 127 \\
&  &  & beam        & 0   & 2   & 0  & 0  & 0  & 119 & 0  & 0 & 121 \\
\bottomrule
    \end{tabular}
\caption{Attribute realization errors on the E2E test set. The Slug model and our delexicalized models delexicalize the NAME and NEAR slots, 
    thus making 0 errors on these attributes. DANGNT and TUDA models perform complete delexicalization. } 
\label{table:autosem}
\end{table*}

Surprisingly, \basegen~using greedy decoding surpases all of the 
baseline systems. This is quite shocking as the Slug model ensembles
three different sequence-to-sequence models producing 10 outputs each using beam search and reranking based on slot alignment to select the final generation
output. The \auggen/\ruleclf~model remains competitive with Slug, 
again even using greedy decoding.
The \auggen/\learnedclf~starts underperforming Slug on BLEU score but
remains competitive on ROUGE-L and METEOR again when using greedy decoding.
Overall the augmented training data tends to hurt generation quality.
In this regard, the added noise of the trained classifier exacerbates things
as it reduces quality more than the rule-based filtering. 

In the lexicalized setting,
\basegen~produces lower quality output than the Slug system.
However, the augmented training procedure increases
the quality of the lexicalized \auggen~model which beats Slug on ROUGE-L.

The automatic quality evaluations are somewhat limited, however. To gain
more insight into model performance
we apply our rule based parser to estimate attribute realization error
for all system outputs on the test set,
similarly to \cite{duvsek2019evaluating}
(e.g., if the MR specifies \textit{food[French]},
we check to make sure the generated utterance says so). 
The results of this evaluation are shown in \autoref{table:autosem}.
Immediately, it is revealed that \basegen~is far worse than the baseline 
methods making 115 and 83 errors using greedy and beam decoding respectively.

The \auggen/\ruleclf~model achieves zero test set
errors even when using the greedy 
decoding. The \auggen/\learnedclf~model is slightly worse (in agreement 
with the automatic quality measurements), but its greedy search is still
superior to the more sophisticated Slug decoder, achieving 19 total
test set errors compared to Slug's 67 errors.

The lexicalized \basegen~model has especially high error rates, 
particularly on the \textit{Name} and \textit{Near} attributes.
With augmented data training, the \auggen~model reduces these errors
 to zero when using greedy search and 2 with beam search. Unfortunately,
the augmented training is more unstable in the lexicalized setting, 
as it produces a large spike in \textit{food} attribute errors, although
the \auggen~models still have lower overall error than  \basegen.

\subsection{Laptops and TVs Self-Training}
We perform similar experiments on the Laptops and TVs datasets. 
We train a separate \basegen~model for each dataset 
for 300 epochs with a learning rate of 0.1
for Laptops and 0.25 for TVs. The weight decay penalty was 0.0001 
and dropout probability was 0.25. Best model iteration is determined
by validation set BLEU score. As in the E2E experiments, we create an augmented
dataset for both the Laptops and TVs dataset using the method
outlined in \autoref{section:selftrain}. We then train new generators 
\auggen~on the union of original training data and the augmented dataset.

\paragraph{Results} 

We automatically evaluate our models using the evaluation script of
\citet{wen2016multi}, which computes BLEU scores, as well as slot
alignment error rate (since this dataset is almost fully delexicalized,
it simply checks for the presence of the correct attribute placeholders
according to the MR). We compare again to the Slug model as well
as the Semantically Conditioned LSTM (SCLSTM) \cite{wen2015semantically}
which report state-of-the-art results on these datasets.

The results are more mixed here. Our BLEU scores are about 15 points 
below the baselines on the Laptops dataset and 20 points below the 
baselines on the TVs dataset. 
Upon examing  the evaluation script in detail we see that 
BLEU score is calculated using 5 model outputs which \citet{juraskaslug2slug}
and \citet{wen2016multi} do. We only produce the 1-best output
at test time,
perhaps explaining the difference.

Looking through our model
outputs we see mostly good utterances, often nearly exactly matching the 
references.
Our models outperform the state of the art models on errors. The best state of the art models  make errors by generating sentences that do not match the input representation 0.79\%  and 1.67\% of the time on the Laptops and TVs datasets
respectively. Our \auggen~model reduces that error to only 0.13\% and 
0.20\%.

\begin{table}
    \begin{tabular}{lrrrr}
        \toprule
        & \multicolumn{2}{c}{Laptops} & \multicolumn{2}{c}{TVs} \\
        Model & BLEU & Err. & BLEU & Err.  \\
        \midrule
        SCLSTM    & 51.16 &   0.79\% & \textbf{52.65} &   2.31\%\\
        Slug & \textbf{52.38}  &  1.55\% & 52.26  &  1.67\% \\
        \basegen~~beam  &37.13  &0.72\% & 32.63 & 0.72\% \\
        \auggen~~greedy & 37.21 &  \textbf{0.13\%} & 32.43 & 0.28\%\\
        ~~~~~~beam   & 37.19 & 0.14\% & 32.59 &  \textbf{0.20\%} \\
\bottomrule
    \end{tabular}

    \caption{BLEU and automatic attribute error on the Laptops and TVs
    datasets.}
    \label{table:laptoptvautoqual}
\end{table}

\subsection{Experiment 4: Human Evaluation} 

\paragraph{E2E Dataset} We had two undergraduate students not involved with 
the research look at 100 random test set utterances for six
of our model variants. They were shown
both the Slug output and one of our model outputs and asked to select
which output was of better linguistic quality and correctness or 
indicate that they were equally good.
 We resolved disagreements in favor of the baseline,
i.e. if any annotator thought the baseline was better we considered it so.
If an annotator marked one of our 
systems as better and the other marked it as equal, we considered it 
equal to the baseline. Inter-annotator agreement was high, with 92\% agreement on correctness
and 88\% agreement on quality.

\autoref{humane2e} shows the results of the evaluation.
We find that the \auggen~model outputs are indistinguishable from the Slug
model in terms of linguistic quality, regardless of the setting.
In terms of correctness, the lexicalized \auggen~model is as good as or better than the Slug model 98\%
of the time. 
When using the delexicalized models, we don't even need beam search.
The delexicalized \auggen~greedy decoder is as good as or better 
than Slug 100\% of the time.

\paragraph{Laptops Dataset} We had the same annotators look at 100 random 
\textit{Inform}
DAs from the Laptops test set since they are the majority DA type and we 
could use the same annotator
guidelines from the E2E experiment. We do not have access to the Slug
or SCLSTM outputs on this dataset, so we compared to one of the two
test set reference sentences (picking at random) vs. 
the \auggen/\ruleclf~with greedy decoding. \autoref{humanlaptop}
shows the results. Despite the low BLEU scores, we find our outputs
to be of comparable quality to references 91\% of the time. Moreover,
they are equally as correct as the human references 100\% of the time.
Annotators agreed 99\% and 87\% of the time on correctness and quality 
respectively.

\begin{table}
\center
    \begin{tabular}{r ccc| ccc}
\toprule
        & \multicolumn{3}{c}{Correct.} & \multicolumn{3}{c}{Quality} \\
    Model & $>$ & $=$ & $<$ & $>$ & $=$ & $<$ \\
\midrule
delex.~\basegen~b & 7 & 89 & 4 & 1 & 96 & 3 \\
delex.~\auggen~\ruleclf~g & 7 & 93 & 0 & 0 & 100 & 0 \\
delex.~\auggen~\ruleclf~b & 7 & 93 & 0 & 0 & 100 & 0 \\
delex.~\auggen~\learnedclf~g & 5 & 95 & 0 & 0 & 100 & 0 \\
delex.~\auggen~\learnedclf~b & 8 & 92 & 0 & 0 & 100 & 0 \\
lex.~\auggen~\ruleclf~g & 8 & 90 & 2 & 0 & 100 & 0 \\
\bottomrule
\end{tabular}
\caption{Human correctness and quality judgments (\%). Comparisons are better
than ($>$), equal to ($=$), and worse than ($<$) the baseline Slug model.
(g) and (b) indicate greedy and beam decoding respectively.}
\label{humane2e}
\end{table}

\begin{table}
    \center
    \begin{tabular}{c ccc| ccc}
\toprule
        & \multicolumn{3}{c}{Correct.} & \multicolumn{3}{c}{Quality} \\
    Model & $>$ & $=$ & $<$ & $>$ & $=$ & $<$ \\
\midrule
delex. \auggen~\ruleclf~g & 0 & 100 & 0 & 2 & 91 & 7 \\

\bottomrule
\end{tabular}
\caption{Human correctness and quality judgments (\%). Comparisons are better
than ($>$), equal to ($=$), and worse than ($<$) the test set references.}
\label{humanlaptop}
\end{table}

\section{Sample Analysis and Discussion}
We hypothesize that self-training improves the correctness of outputs
by sacrificing some expressivity of the model. For example, 
\auggen~BLEU scores
on the E2E dataset drop by at least 0.8 as compared to \basegen~with beam
search. We see a similar pattern on the TVs dataset. Self-training
increases automatic metrics in the lexicalized setting, but this could 
be attributable to reductions  in \textit{Name} and \textit{Near}
realization errors, which are orthogonal to the 
syntactic diversity of generation.

To better quantify these effects we report the average length in words, 
average number of sentences, and average revised Developmental Level (D-Level)
score according to the D-Level analyser \cite{lu2009automatic}.
The D-Level analyser automatically categorizes the syntactic complexities 
of an utterance into one of eight categories, with eight being the most
complex, based on the revised Developmental Level scale \cite{rosenberg1987indicators,covington2006complex}.

\autoref{table:samplequal} shows the statistics for the E2E test set outputs. 
In the lexicalized setting, the mean D-level results support our hypothesis;
syntactic complexity of test set outputs decreases from \basegen~to~\auggen.
In the delexicalized setting this is somewhat true; three of 
the \auggen~models have lower mean D-level scores than \basegen~with 
greedy decoding. Curiously, \auggen/\learnedclf~with beam search has the 
highest overall syntactic complexity of any our model variants, at odds
with our hypothesis.
No models are as syntactically complex as the human references, but our
models come closest, with a mean D-Level category of 1.87 using the delex. \auggen/\learnedclf~model with beam decoder. 

We  also 
see that \auggen/\ruleclf~models are over two sentences
in length on average while the human references are under two sentences,
suggesting they are more often falling back to simple but reliable
ways to realize attributes (e.g., appending ``It is a family-friendly venue.'').

\begin{table}[t]
    \begin{tabular}{llll}
    \toprule
    Model & Words & Sents & MDL \\
    \midrule
    Human Refs. & 24.06& 1.76& \textbf{2.25}\\
    Slug &24.20 &1.86 & 1.39 \\
    \midrule
    lex. \basegen~greedy &25.73 &2.18 & \textbf{1.84} \\
    lex. \basegen~beam   &26.00 &2.20 & 1.50 \\
    lex. \auggen~\ruleclf~greedy & 26.01& 2.20& 1.39 \\
    lex. \auggen~\ruleclf~beam   & 26.04&2.17 & 1.45 \\
    \midrule
    delex. \basegen~greedy & 24.83& 2.10 & 1.79 \\
    delex. \basegen~beam &24.51 & 2.03 & 1.48 \\
    delex. \auggen~\ruleclf~greedy & 26.50 & 2.29 & 1.74 \\
    delex. \auggen~\ruleclf~beam & 26.46 & 2.28 & 1.74 \\
    delex. \auggen~\learnedclf~greedy & 25.33&1.76 & 1.77 \\
    delex. \auggen~\learnedclf~beam &25.49 &1.75 & \textbf{1.87} \\
    \bottomrule
    \end{tabular}
    \caption{Words/sentences per utterance
    and mean D-Level score of model outputs on the E2E dataset.}
\end{table}

\begin{table}
\center
\setlength{\tabcolsep}{4pt}
\begin{tabular}{crrrr}
\toprule
\augdata & Size & Words & Sents & MDL\\
\midrule
delex. $\augdata_{\learnedclf}$  & 384,436 & 22.5 & 2.0 & 1.77 \\
delex. $\augdata_{\ruleclf}$ & 501,909 & 22.7 & 2.1 & 1.76 \\
lex. $\augdata_{\ruleclf}$ & 1,591,778 & 23.2 & 2.1 & 1.69 \\
\bottomrule
\end{tabular}
\caption{E2E augmented dataset statistics: total utterances, 
words per utterance,
sentences per utterance, and mean D-Level score.}
\label{table:samplequal}
\end{table}

That our simple models with greedy search and no semantic control mechanisms
can perform as reliably as more sophisticated models suggest that 
in standard training regimes we 
are often not fully learning from all information available in the 
training data. Via sampling we can uncover novel recombinations of 
utterances that are only implied by the provided references.
The gains of self-training also suggest that additional
research into active learning for this task might bear fruit.

One curious observation about the self-training procedure is that 
it leads to a convergence in output complexity of greedy and beam decoding.
The differences between mean D-level score on the \basegen~models is
0.34 and 0.31 in the lexicalized and delexicalized settings respectively.
This shrinks to 0.0 and 0.1 in the delexicalized \auggen~settings and 0.06 
for lexicalized \auggen, suggesting that the model probability distributions
are sharpening around a smaller set of output structures.

\section{Related Work}
Neural encoder-decoder
models, are a popular choice for dialog generation \cite{mei2015talk,duvsek2016sequence,chen2018sheffield,juraskaslug2slug,elder2018e2e}. 
However, the quality can vary significantly, with relatively
similar architectures yielding both poor and competitve performance
 \cite{duvsek2019evaluating}. 
All of the cited work on the E2E or Laptops and TVs datasets 
uses beam search to achieve competitive performance.
In addition, they often employ reranking to ensure that all attributes
are realized \cite{duvsek2016sequence,juraskaslug2slug,wen2015semantically}.
\cite{elder2018e2e} employ
pointer generators to directly copy attribute values, while also 
using coverage penalties on the attention weights to ensure that 
all attribute slots are attended to. Unlike these approaches, we do not require
beam search, reranking, or other specialized attention mechanisms or loss
functions to obtain low error rates. Instead we use data-augmentation
to obtain a more reliable but simpler model.

Data augmentation has also been used by prior neural generation models.
\citet{juraskaslug2slug} breaks multi-sentence utterances into separate training
instances. They also try training on more complex sentences alone but this
model was less reliably able to realize all attributes correctly.
They also  
do not generate new utterance/MR pairs for training as we do.

Our method is in some ways similar to the reconstructor setting of 
\citet{shen2019pragmatically}, where a base speaker model $S_0$ produces
utterances and a listener model $L$ reconstructs the input MR.
In the framework of rational speech acts (RSA) \cite{monroe2015learning},
a rational speaker model is obtained by composing the base speaker and listener,
i.e. $S_1(y|x) = L(x|y)\cdot S_0(y|x)$.
While we do not directly compose our parser \clf~and \basegen, the \auggen~model
is learning from the composition of the two.
The theoretical commitments  of RSA are somewhat orthogonal to our approach.
It would be interesing to combine both methods,
by incorporating self-training into the RSA framework.

\section{Conclusion}
We present a novel self-training methodology for learning DNN-based dialogue
generation models using noise injection sampling and a MR parser. 
Even with relatively simple architectures and greedy decoding we are able 
to match the performance of state-of-the-art baselines on automatic measures 
of  quality while also achieving superior semantic correctness. These 
findings hold under a human evaluation as well. On automatic measures
of syntactic complexity we also find our approach is closer to matching
human authored references than prior work. 
In  future work, we intend
to explore methods of self-training that futher improve syntactic
diversity.

\section{Acknowledgements}
The authors would like to thank the anonymous reviewers for their valuable feedback, as well as Thomas Effland, Katy Ilonka Gero, and Christopher Hidey for their comments during the drafting of this paper.

This research is based upon work supported in
part by the Office of the Director of National Intelligence (ODNI), Intelligence Advanced Research
Projects Activity (IARPA), via contract \#FA8650-17-C-9117. The views and conclusions contained
herein are those of the authors and should not be
interpreted as necessarily representing the official
policies, either expressed or implied, of ODNI,
IARPA, or the U.S. Government. The U.S. Government is authorized to reproduce and distribute
reprints for governmental purposes notwithstanding any copyright annotation therein.

\bibliography{acl2019}
\bibliographystyle{acl_natbib}

\clearpage 
\appendix
\begin{figure}[h]
    \center
\begin{tabular}{ll}
\toprule
\textbf{Compare DA} \\
~\\
\multicolumn{2}{c}{
    \small{
    $\textit{Compare}\left(\begin{array}{ll}
            \textit{name[Atlas 89]} & \textit{name[Eurus 93]} \\ 
            \textit{isForBusiness[yes]} & \textit{isForBusiness[no]} \\
            \textit{weight[2.3]} & \textit{weight[1.12]}\\
        \end{array}\right)
$}}\\
~\\ 
\textbf{InformCount DA} \\
~\\
\multicolumn{2}{c}{
    \small{
    $\textit{InformCount}\left(\begin{array}{l}
            \textit{count[40]} \\
        \textit{family[don't~care]}\\
        \textit{batteryRating[excellent]}\\
    \end{array}
    \right)$}
}\\

~\\
\textbf{Compare Input} & \textbf{InformCount Input}\\
\midrule
\small{
$\begin{array}{ll}
\mrx_1 = & \textrm{compare}\\
\mrx_2 = & \textrm{name} \\
\mrx_3 = & \textrm{is\_for\_biz\_yes} \\
\mrx_4 = & \textrm{weight} \\
\mrx_5 = & \textrm{name} \\
\mrx_6 = & \textrm{is\_for\_biz\_no} \\
\mrx_7 = & \textrm{weight} \\
\end{array}$
}
&
\small{
$\begin{array}{l}
 \textrm{inform\_count}\\
 \textrm{count} \\
 \textrm{family\_dont\_care} \\
 \textrm{batteryrating} \\
\textrm{---} \\
\textrm{---} \\
\textrm{---} \\
\end{array}$
}
\\
                   ~\\
\textbf{Compare Utterance}\\
\midrule
\multicolumn{2}{p{7cm}}{

weighing $\textrm{WEIGHT}_1$ kg for business computing the $\textrm{NAME}_1$ 
is compared to the $\textrm{NAME}_2$ which weighs $\textrm{WEIGHT}_2$ kg and 
is not for business computing . which one do you like
} \\
~\\
~\\
\textbf{InformCount Utterance}\\
\midrule
\multicolumn{2}{p{7cm}}{
there are COUNT laptop -s with an BATTERYRATING battery rating if you do not care about the product family.
}\\
\bottomrule
\end{tabular}
~\\
\caption{Example MR, input representations, and utterances for the Laptops 
    and TVs datasets.
  }
\label{figure:laptoptvexample}
\end{figure}

\section{Additional Preprocessing Details}
\label{app:preproc}

MR attributes can be one 
of two types, dictionary attributes, e.g. \textit{Name} or
\textit{CustomerRating}, where the value for the attribute comes from a 
closed set of valid slot fillers, and binary attributes, e.g. \textit{familyFriendly} or \textit{hasUsbPort}, which can have values \textit{yes} or 
\textit{no}.
Additionally, on the Laptops and TVs datasets, attributes can also 
have a distinguished \textit{don't care} value, e.g. the MR
\textit{InformCount$\big($count[40], priceRange[don't~care]$\big)$} could be realized
as ``There are 40 laptops available if you do not care about the price range.''

When using the delexicalied model, we do not represent
the \textit{Name} attribute in the input \mrx, since every valid E2E MR contains
this attribute, i.e. the model learns on its own to always generate the 
\textit{NAME} token to be replaced at test time. 

Certain attributes were inconsistently realized in the reference utterances.
E.g., utterances would frequently refer to locations as restaurants even when
the \textit{eatType} attribute was not present
so we ammended the MR to have the attribute \textit{eatType[restaurant]}
in those cases.
Additionally, \textit{priceRange[cheap]} and \textit{priceRange[less than 
\pounds{}20]} would be interchanged; \textit{cheap} utterances that mentioned
numerical amounts were remapped to \textit{less than \pounds{}20} and visa-versa. Similar
corrections were made for the \textit{customerRating} attribute.
Note, these changes were only done on the training and validation 
set. We do not modify the test set at all.

\section{Model Input Representation}
\label{app:inputrep}

\begin{figure}
\begin{tabular}{p{3.5cm} p{3.2cm}}
\toprule
\textbf{Inform DA}\\
\multicolumn{2}{c}{
\small{
    $\textit{Inform}\left(\begin{array}{l}
            \textit{name[The~Mill]} \\
        \textit{near[Avalon]}\\
        \textit{food[Italian]}\\
    \end{array}
    \right)$}
}\\

~\\
\textbf{Input Lexicalized} & \textbf{Input Delexicalized}\\
\midrule
\small{
$\begin{array}{ll}
\mrx_1 & = \textrm{eat\_type\_n/a} \\ 
\mrx_2 & = \textrm{near\_avalon} \\
\mrx_3 & = \textrm{area\_n/a} \\
\mrx_4 & = \textrm{fam.\_friend.\_n/a} \\
\mrx_5 & = \textrm{cust.\_rating\_n/a} \\
\mrx_6 & = \textrm{price\_range\_n/a} \\
\mrx_7 & = \textrm{food\_Italian} \\
\mrx_8 & = \textrm{name\_the\_mill} \\
\end{array}$
}&
\small{
$\begin{array}{l}
 \textrm{eat\_type\_n/a} \\ 
 \textrm{near\_present} \\
 \textrm{area\_n/a} \\
 \textrm{fam.\_friend.\_n/a} \\
 \textrm{cust.\_rating\_n/a} \\
 \textrm{price\_range\_n/a} \\
 \textrm{food\_Italian} \\
 \textrm{ ---} \\
\end{array}$}
 \\
~\\
\multicolumn{2}{l}{\textbf{Lexicalized Utterance}} \\
\midrule
\multicolumn{2}{l}{
the mill serves up italian food near avalon . }\\
~\\
\multicolumn{2}{l}{\textbf{Delexicalized Utterance}} \\
\midrule
\multicolumn{2}{l}{
NAME serves up italian food near NEAR .} \\
\bottomrule
\end{tabular}
\caption{Example MR, input representation, and utterance for the E2E dataset.}
\label{figure:e2einp}
\end{figure}

We show an example input sequence \mrx~for the E2E dataset in \autoref{figure:e2einp} and for the Laptops and TVs datasets in \autoref{figure:laptoptvexample}.

\section{CNN Classifier Details}
\label{app:cnnclf}

We use a separate CNN classifier for each attribute to predict
the corresponding value (or \textit{n/a}) from an utterance \descy.
We first look up the tokens in \descy~ in an embedding matrix $E$
to obtain a matrix $W\in\mathbb{R}^{N \times D}$ where $D=50$ is
the embedding dimension.

We then apply a series of unigram, bigram, and trigram convolutional
filters each with $50$ output features.
After concatenating and max-pooling over the sequence dimension,
and applying a ReLU activation,
we obtain a hidden layer in $\mathbb{R}^{150}$.
We then apply another fully-connected layer with ReLU activation
which down projects the hidden layer to $\mathbf{R}^{50}$.
Finaly we apply the final softmax layer to predict the class label.

During training we apply dropout (with drop rate 0.25) to 
the embedding layer, convolutional filter outputs, and hidden
layers. We train for 30 epochs with gradient descent
using  a learning rate of 0.25 and 
weight decay penalty of 0.0001, using validation set F1
as our model selection criterion.

We treat the utterance as a

\section{Augmented Dataset Samples}
\label{app:samples}

Samples and their parsed MR for the E2E datsets are shown in 
\autoref{table:delex:rules:e2e:sample:example},
\autoref{table:delex:clf:e2e:sample:example}, and
\autoref{table:lex:rules:e2e:sample:example}.

\begin{table*}\center\begin{tabular}{p{15.4cm}}\toprule
\textbf{Inform}(\textit{eatType}[pub] \textit{food}[Italian] \textit{name}[NAME] \textit{priceRange}[high]) \\
NAME is a pub that serves italian food in the high price range . \\
~\\
\textbf{Inform}(\textit{area}[city~centre] \textit{customerRating}[5~out~of~5] \textit{eatType}[coffee~shop] \textit{familyFriendly}[no] \textit{food}[Indian] \textit{name}[NAME]) \\
the NAME is a coffee shop that serves indian food . it is located in the city centre and has a customer rating of 5 out of 5 . it is not kid friendly . \\
~\\
\textbf{Inform}(\textit{customerRating}[3~out~of~5] \textit{eatType}[pub] \textit{food}[English] \textit{name}[NAME] \textit{near}[NEAR]) \\
the NAME is a pub that serves english food and is located near the NEAR . it has a customer rating of 3 out of 5 . \\
~\\
\textbf{Inform}(\textit{area}[city~centre] \textit{customerRating}[3~out~of~5] \textit{eatType}[pub] \textit{familyFriendly}[yes] \textit{food}[Fast~food] \textit{name}[NAME] \textit{priceRange}[less~than~\pounds20]) \\
the NAME is a pub in the city centre that serves fast food for less than \pounds 20 . it has a customer rating of 3 out of 5 and is family - friendly . \\
~\\
\textbf{Inform}(\textit{customerRating}[high] \textit{eatType}[coffee~shop] \textit{food}[French] \textit{name}[NAME] \textit{near}[NEAR] \textit{priceRange}[cheap]) \\
NAME is a cheap coffee shop near NEAR that serves french food . it has a high customer rating . \\
~\\
\textbf{Inform}(\textit{area}[city~centre] \textit{customerRating}[5~out~of~5] \textit{eatType}[pub] \textit{familyFriendly}[yes] \textit{food}[Chinese] \textit{name}[NAME] \textit{priceRange}[cheap]) \\
the NAME is a family - friendly pub in the city centre . it serves cheap chinese food and has a customer rating of 5 out of 5 . \\
~\\
\textbf{Inform}(\textit{customerRating}[high] \textit{familyFriendly}[yes] \textit{food}[Indian] \textit{name}[NAME] \textit{priceRange}[high]) \\
the NAME serves high priced indian food . it has a high customer rating and is child friendly . \\
~\\
\textbf{Inform}(\textit{customerRating}[3~out~of~5] \textit{food}[Japanese] \textit{name}[NAME] \textit{near}[NEAR] \textit{priceRange}[less~than~\pounds20]) \\
NAME serves japanese food for less than \pounds 20 . it is located near NEAR and has a customer rating of 3 out of 5 . \\
~\\
\bottomrule\end{tabular}
\caption{Example E2E samples obtained using noise injection sampling with delex. \basegen~and \ruleclf~to parse the MR.}
\label{table:delex:rules:e2e:sample:example}
\end{table*}

\begin{table*}\center\begin{tabular}{p{15.4cm}}\toprule
\textbf{Inform}(\textit{customerRating}[3~out~of~5] \textit{familyFriendly}[no] \textit{food}[English] \textit{name}[NAME] \textit{priceRange}[\pounds20-25]) \\
the NAME serves english food in the \pounds 20 - \pounds 25 price range . it is not kid friendly and has a customer rating of 3 out of 5 \\
~\\
\textbf{Inform}(\textit{area}[riverside] \textit{customerRating}[3~out~of~5] \textit{eatType}[restaurant] \textit{familyFriendly}[yes] \textit{food}[Chinese] \textit{name}[NAME] \textit{near}[PRESENT] \textit{priceRange}[high]) \\
NAME is a family friendly restaurant serving chinese food in the riverside area near NEAR . it has a high price range and a customer rating of 3 out of 5 \\
~\\
\textbf{Inform}(\textit{area}[city~centre] \textit{eatType}[restaurant] \textit{food}[English] \textit{name}[NAME] \textit{priceRange}[less~than~\pounds20]) \\
NAME is a restaurant providing english food in the less than \pounds 20 price range . it is located in the city centre \\
~\\
\textbf{Inform}(\textit{customerRating}[3~out~of~5] \textit{food}[Italian] \textit{name}[NAME] \textit{near}[PRESENT] \textit{priceRange}[\pounds20-25]) \\
NAME has a price range of \pounds 20 - 25 . it has a customer rating of 3 out of 5 and serves italian food . it is near the NEAR \\
~\\
\textbf{Inform}(\textit{area}[riverside] \textit{customerRating}[3~out~of~5] \textit{food}[Chinese] \textit{name}[NAME] \textit{priceRange}[moderate]) \\
the NAME serves chinese food in the riverside area . it has a moderate price range and a customer rating of 3 out of 5 \\
~\\
\textbf{Inform}(\textit{area}[riverside] \textit{customerRating}[low] \textit{familyFriendly}[yes] \textit{food}[English] \textit{name}[NAME]) \\
the NAME serves english food in the riverside area . it has a low customer rating and is kid friendly \\
~\\
\textbf{Inform}(\textit{customerRating}[low] \textit{eatType}[coffee~shop] \textit{familyFriendly}[yes] \textit{food}[Japanese] \textit{name}[NAME]) \\
the NAME is a family friendly japanese coffee shop with a low customer rating \\
~\\
\textbf{Inform}(\textit{customerRating}[1~out~of~5] \textit{familyFriendly}[no] \textit{food}[Japanese] \textit{name}[NAME] \textit{near}[PRESENT]) \\
NAME serves japanese food near NEAR . it is not kid friendly and has a customer rating of 1 out of 5 \\
~\\
\bottomrule\end{tabular}
\caption{Example E2E samples obtained using noise injection sampling with delex. \basegen~and \learnedclf~to parse the MR.}
\label{table:delex:clf:e2e:sample:example}
\end{table*}

\begin{table*}\center\begin{tabular}{p{15.4cm}}\toprule
\textbf{Inform}(\textit{area}[riverside] \textit{customerRating}[5~out~of~5] \textit{eatType}[restaurant] \textit{food}[French] \textit{name}[The~Eagle] \textit{near}[The~Sorrento] \textit{priceRange}[less~than~\pounds20]) \\
the eagle is a french restaurant with a 5 out of 5 customer rating and a price range of less than \pounds 20 . it is located in the riverside area near the sorrento . \\
~\\
\textbf{Inform}(\textit{area}[city~centre] \textit{eatType}[coffee~shop] \textit{familyFriendly}[no] \textit{food}[English] \textit{name}[The~Wrestlers] \textit{near}[Raja~Indian~Cuisine]) \\
the wrestlers is a coffee shop that serves english food . it is located in the city centre near raja indian cuisine . it is not child friendly . \\
~\\
\textbf{Inform}(\textit{area}[riverside] \textit{name}[Taste~of~Cambridge] \textit{priceRange}[cheap]) \\
taste of cambridge is located in the riverside area . it is cheap . \\
~\\
\textbf{Inform}(\textit{customerRating}[3~out~of~5] \textit{eatType}[restaurant] \textit{familyFriendly}[yes] \textit{name}[Zizzi]) \\
there is a kid friendly restaurant called zizzi . it has a customer rating of 3 out of 5 . \\
~\\
\textbf{Inform}(\textit{area}[city~centre] \textit{eatType}[pub] \textit{name}[The~Cambridge~Blue] \textit{near}[Yippee~Noodle~Bar] \textit{priceRange}[high]) \\
the cambridge blue is a pub in the high price range . it is located in the city centre near the yippee noodle bar . \\
~\\
\textbf{Inform}(\textit{area}[riverside] \textit{customerRating}[average] \textit{name}[The~Phoenix]) \\
the phoenix is located in the riverside area near the riverside . it has an average customer rating . \\
~\\
\textbf{Inform}(\textit{customerRating}[average] \textit{familyFriendly}[yes] \textit{food}[Indian] \textit{name}[Loch~Fyne]) \\
loch fyne provides indian food . it is family friendly and has an average customer rating . \\
~\\
\textbf{Inform}(\textit{area}[riverside] \textit{customerRating}[1~out~of~5] \textit{food}[Italian] \textit{name}[The~Phoenix] \textit{near}[The~Six~Bells]) \\
the phoenix is located in the riverside area near the city centre , near the six bells . it serves italian food and has a customer rating of 1 out of 5 . \\
~\\
\bottomrule\end{tabular}
\caption{Example E2E samples obtained using noise injection sampling with lex. \basegen~and \ruleclf~to parse the MR.}
\label{table:lex:rules:e2e:sample:example}
\end{table*}

\end{document}